\documentclass[conference]{IEEEtran}
\IEEEoverridecommandlockouts
\usepackage{cite}
\usepackage{amsmath,amssymb,amsfonts}
\usepackage[table, x11names]{xcolor}
\usepackage{algorithmic}
\usepackage{graphicx}
\usepackage{subfigure}
\usepackage{booktabs}
\usepackage{multirow}
\usepackage{fancyhdr}
\usepackage{hyperref}
\usepackage{listings} 
\usepackage{changepage} 
\usepackage[figurename=Fig., tablename=Tab., small]{caption}[2008/04/01]
\def\eg{\textit{e.g}. } 
\def\ie{\textit{i.e}. } 
\def\cf{\textit{cf}. }

\def\wrt{w.r.t. } 

\def\etal{\textit{et al}. }
\usepackage{textcomp}
\usepackage{xcolor}
\def\BibTeX{{\rm B\kern-.05em{\sc i\kern-.025em b}\kern-.08em
    T\kern-.1667em\lower.7ex\hbox{E}\kern-.125emX}}
\begin{document}

\title{Reliable Detection of Doppelgängers based on\\ Deep Face Representations}

\author{\IEEEauthorblockN{C. Rathgeb, D. Fischer, P. Drozdowski, C. Busch}
\IEEEauthorblockA{\textit{da/sec -- Biometrics and Internet Security Research Group} \\
 Hochschule Darmstadt, Germany \\
\texttt{\{christian.rathgeb,daniel.fischer,christoph.busch\}@h-da.de}}
}

\maketitle

\begin{abstract}
Doppelgängers (or lookalikes) usually yield an increased probability of false matches in a facial recognition system, as opposed to random face image pairs selected for non-mated comparison trials. In this work, we assess the impact of doppelgängers on the HDA Doppelgänger and  Disguised Faces in The Wild  databases using a state-of-the-art face recognition system. It is found that doppelgänger image pairs yield very high similarity scores resulting in a significant increase of false match rates. Further, we propose a doppelgänger detection method which distinguishes doppelgängers from mated comparison trials by analysing differences in deep representations obtained from face image pairs. The proposed detection system employs a machine learning-based classifier, which is trained with generated doppelgänger image pairs utilising face morphing techniques. Experimental evaluations conducted on the  HDA Doppelgänger  and  Look-Alike Face databases reveal a detection equal error rate of approximately 2.7\% for the task of separating mated authentication attempts from doppelgängers.
\end{abstract}
\vspace{0.2cm}
\vspace{0.1cm}

\section{Introduction}\label{sec:introduction}
Face recognition technologies are used in many personal, commercial, and governmental identity management systems worldwide. Developments in convolutional neural networks have achieved remarkable improvements in facial recognition accuracy, surpassing human-level performance \cite{Guo-DeepFaceSurvey-2019,Taigman14,Ranjan18a}. In particular, state-of-the-art deep face recognition systems turn out to be robust against a variety of covariates, such as variations in  expression \cite{9039580}, ageing \cite{Best-Rowden18}, beautification \cite{Rathgeb-ImpactDetectionFacialBeautificationSurvey-ACCESS-2019}, or poor sample quality \cite{Schlett}.

The improved robustness of facial recognition systems reduces the number of false rejections, however, it may also increase the vulnerability to impostors. This has for instance been shown for presentation attacks where an attacker aims at impersonating a target subject by using some presentation attack instrument \cite{Mohammadi2018}. In contrast to specific attacks, in a \emph{zero-effort impostor} attempt, an individual submits their own biometric characteristic while attempting to obtain a successful authentication against another subject \cite{ISO-IEC-19795-1:2021}. It is important to note that a zero-effort impostor trial do \textit{not} necessarily represent an attack. Previous works reported high chances of false matches in zero-effort impostor attempts in the presence of kin-relationship, in particular for monozygotic, \ie identical, twins \cite{Pruitt11,Phillips11a}. This effect is far less pronounced for other popular biometric characteristics, \eg fingerprint \cite{Tao12} or iris \cite{Daugman20}. 

\begin{figure}[!t]
\vspace{0.1cm}
\centering
\includegraphics[height= 3.cm]{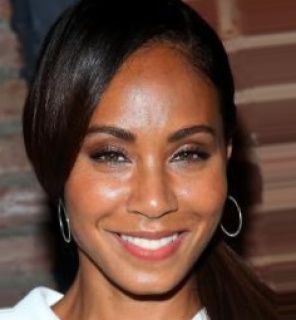} \includegraphics[height= 3.cm]{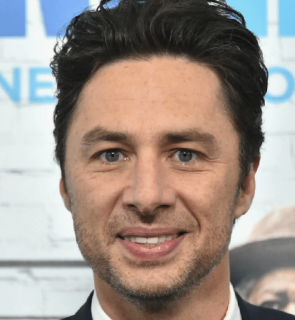} \vspace{0.1cm} \includegraphics[height= 3.cm]{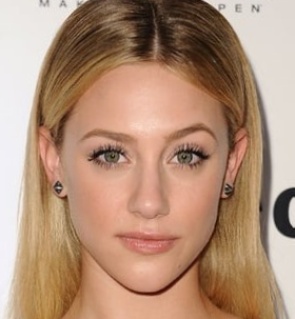}
\includegraphics[height= 3.cm]{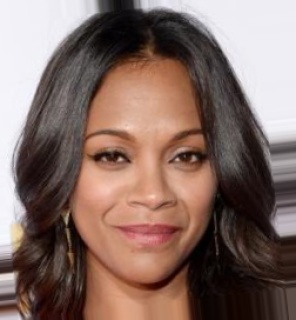} \includegraphics[height= 3.cm]{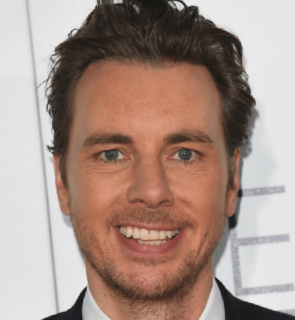} \includegraphics[height= 3.cm]{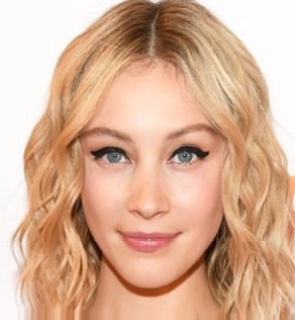}
\vspace{0.1cm}
\caption{Example doppelgänger image pairs (column-wise) from the HDA Doppelgänger database.}\label{fig:ex}\vspace{-0.4cm}
\end{figure}

In contrast to monozygotic twins, so-called \emph{doppelgängers} refer to biologically unrelated lookalikes, which have been reported to have high chances of false matches in zero-effort impostor attempts, too \cite{Rathgeb-Doppelgaenger-BIOSIG-2021}. Examples are depicted in figure~\ref{fig:ex}. Apart from demographic attributes, doppelgängers often share facial properties such as facial shape. Additionally, some facial properties may further be altered to obtain even higher similarity towards a target subject, \eg by using hair style or the use of makeup \cite{Rathgeb-MakeupAttackDetection-ACCESS-2020}. Like identical twins, doppelgängers are expected to yield high success probabilities as opposed to random zero-effort impostor attempts. This effect has already been confirmed by several researchers, \eg \cite{Lamba11,MOEINI20171,Smirnov17}. This may lead to serious risks in various scenarios, like blacklist checks, where innocent subjects may have a higher chance to match to a lookalike in the list. So far, only a few works have been devoted to detecting doppelgängers in facial recognition systems, \eg \cite{MOEINI20171,Rustam_2019}. However, proposed detection schemes have reported unpractical error rates, \ie above 10\%. This prevents such detection methods from being used in real-world deployments of face recognition, since they would either not be able to reliably detect doppelgängers at low false positive rates or cause an unacceptable amount of false rejections of the overall system. 

\begin{table*}[!t]
\begin{center}
\caption{Overview of most relevant works on the impact and detection of doppelgängers in face recognition systems.}
\label{table:related}
\renewcommand{\arraystretch}{1.1}
\begin{tabular}{cccll}
\toprule
\textbf{Year} & \textbf{Authors} & \textbf{Database} & \textbf{Description} & \textbf{Results}\\
\midrule
2011 & \begin{tabular}{@{}c@{}} Lamba \etal \\\cite{Lamba11} \end{tabular} & \begin{tabular}{@{}c@{}}  Look-Alike Face (LAF) \\database (web-collected), \\  250 image pairs\end{tabular}  & \begin{tabular}{@{}l@{}} Preliminary study on the ability of 50 humans and\\ automated face recognition; \underline{Detection}: Region-\\based feature extraction with SVM. \\ \end{tabular} & \begin{tabular}{@{}l@{}}Close to random performance for automated\\ algorithms and humans (latter increases for\\ familiar faces); \underline{Detection}: $\sim$40\% D-EER \end{tabular} \\\midrule
2017 & \begin{tabular}{@{}c@{}} Moeini \etal \\\cite{MOEINI20171}\end{tabular}& \begin{tabular}{@{}c@{}}LAF \cite{Lamba11} \end{tabular} & \begin{tabular}{@{}l@{}}\underline{Detection}: fusion of texture analysis and 3D re-\\construction with SRC. \end{tabular} & \begin{tabular}{@{}c@{}} \underline{Detection}: $\sim$6\% D-EER \end{tabular} \\\midrule
2017 & \begin{tabular}{@{}c@{}} Smirnov \etal \\\cite{Smirnov17}\end{tabular}& \begin{tabular}{@{}c@{}}One-shot Face\\ Recognition database \end{tabular} & \begin{tabular}{@{}l@{}} Sampling method to identify doppelgängers which\\ are inserted into face representation
learning. \end{tabular} & \begin{tabular}{@{}l@{}} Significant improvement in representation\\ learning for face recognition. \end{tabular} \\\midrule
2017 & \begin{tabular}{@{}c@{}} Deng \etal \\\cite{deng2017fine}\end{tabular}& \begin{tabular}{@{}c@{}} Labeled Faces\\ in the Wild (LFW) \end{tabular} & \begin{tabular}{@{}l@{}}Collection of FGLFW database by  crowd-sourcing\\  (3,000 similarly-looking face pairs) and impact\\ assessment for different face recognition systems. \end{tabular} & \begin{tabular}{@{}c@{}} Drop of recognition accuracy by 10\%-20\%. \end{tabular} \\\midrule
2019 & \begin{tabular}{@{}c@{}} Rustam and \\ Faradina \cite{Rustam_2019}\end{tabular}& \begin{tabular}{@{}c@{}}LAF \cite{Lamba11} \end{tabular} & \begin{tabular}{@{}l@{}} \underline{Detection}: application of Fuzzy Kernel C-Means\\ for identifying lookalikes.\end{tabular} & \begin{tabular}{@{}l@{}}\underline{Detection}: $\sim$25\% D-EER \end{tabular} \\\midrule
2020 & \begin{tabular}{@{}c@{}} Singh \etal \\\cite{Singh19}\end{tabular}& \begin{tabular}{@{}c@{}} Disguised Faces\\ in the Wild (DFW)\\  (web-collected),\\ 4,440 image pairs \end{tabular} & \begin{tabular}{@{}l@{}} Collection of DFW dataset and benchmark of  face\\ recognition systems in a competition with different\\ difficulty levels. \end{tabular} & \begin{tabular}{@{}l@{}}$\sim$55\% GMR at 0.1\% FMR (best algorithms\\ on impersonation subset)\end{tabular} \\\midrule
2021 & \begin{tabular}{@{}c@{}} Swearingen \\  and Ross \cite{Swearingen20}\end{tabular} & \begin{tabular}{@{}c@{}}TinyFace database,\\679K image pairs \end{tabular} & \begin{tabular}{@{}l@{}}Re-ranking of candidate list using a lookalike dis-\\ambiguator trained to distinguish lookalikes. \end{tabular} & \begin{tabular}{@{}l@{}}Modest improvement in 
accuracy in a closed\\-set identification. \end{tabular} \\\midrule
2021 & \begin{tabular}{@{}c@{}} Rathgeb \etal \\\cite{Rathgeb-Doppelgaenger-BIOSIG-2021} \end{tabular}& \begin{tabular}{@{}c@{}}HDA Doppelgänger\\ database (web-collected),\\ 400 image pairs \end{tabular} & \begin{tabular}{@{}l@{}}Collection of doppelgänger image pairs and impact\\ assessment for different face recognition systems.\end{tabular} & \begin{tabular}{@{}l@{}}$\sim$30\% FMR compared to 0.1\% FMR for\\ random impostors. \end{tabular} \\\midrule
\rowcolor{gray!15} 2022 & \begin{tabular}{@{}c@{}} This work \end{tabular}& \begin{tabular}{@{}c@{}}\cite{Singh19,Rathgeb-Doppelgaenger-BIOSIG-2021,Lamba11}\\ \end{tabular} & \begin{tabular}{@{}l@{}}Impact assessment for a face recognition system\\ and detection based on deep face representations.\end{tabular} & \begin{tabular}{@{}l@{}}\underline{Detection}: $\sim$2.7\% D-EER \end{tabular} \\
\bottomrule
\end{tabular}




\end{center}
D-EER: Detection Equal Error Rate, GMR: Genuine Match Rate, FMR: False Match Rate\vspace{-0.2cm}
\end{table*}

A system which reliably distinguishes mated comparison trials from doppelgängers can be integrated in face recognition systems in various ways. Firstly, the detection method could replace a face recognition system, in case the doppelgänger detection method also performs well in separating non-mated comparison trials from mated ones. Actually, this is likely the case, since non-mated comparison trials are expected to yield lower comparison scores, compared to doppelgängers. Secondly, the doppelgänger detection method could be applied in a second line to comparison trails which resulted in a match in the face recognition system. Independent of how a doppelgänger detection method is integrated to a face recognition pipeline, it needs to be specified how to proceed. In an operational scenario, \eg automated border control, the decision of the doppelgänger detection method could be passed to a human examiner (human-in-the-loop system) who could then manually examine the comparison trial. A human examiner's decision could rely on the analysis of further features which are normally not processed in an automated face recognition system, \eg ear shape or facial marks.

In this work, we employ standardised methodologies and metrics \cite{ISO-IEC-30107-3-PAD-metrics-170227} to evaluate the vulnerability of a Commercial Off-The-Shelf (COTS) face recognition system against doppelgängers contained in the HDA Doppelgänger Face Database\footnote{
HDA Doppelgänger Face Database:\newline \url{https://dasec.h-da.de/research/biometrics/hda-doppelgaenger-face-database/}} (hereafter referred to as HDA-DG) \cite{Rathgeb-Doppelgaenger-BIOSIG-2021} and a subset of the Disguised Faces in The Wild  dataset \cite{Singh19}.  These databases have been shown to be extremely challenging for state-of-the-art face recognition systems.  It is found that the doppelgänger image pairs of the HDA-DG database yield very high similarity scores resulting in a significant increase of false match rates. Further, a doppelgänger detection system is introduced. In this system, deep face representations are estimated from a pair of face images employing a state-of-the-art face recognition system. Detection scores are obtained from machine learning-based classifiers analysing differences in deep face representations. Said classifiers are trained with a  database of doppelgängers (and mated authentication attempts) which are generated using face morphing techniques. In experiments conducted on the HDA-DG and Look-Alike Face (LAF) database, a detection equal error rate of $\sim$2.7\% is obtained which significantly outperforms results reported in previously published approaches.

The remainder of this work is organised as follows:  section~\ref{sec:related} summarises relevant related works. The proposed doppelgänger detection method is described in section~\ref{sec:system}. Subsequently, the experimental setup is summarised in section~\ref{sec:db} and results of the impact assessment and detection are presented in section~\ref{sec:experiments}. Finally, conclusions are drawn in section~\ref{sec:conclusion}.
\begin{figure*}[!t]

\vspace{-0.0cm}
\centering
\includegraphics[width=\linewidth]{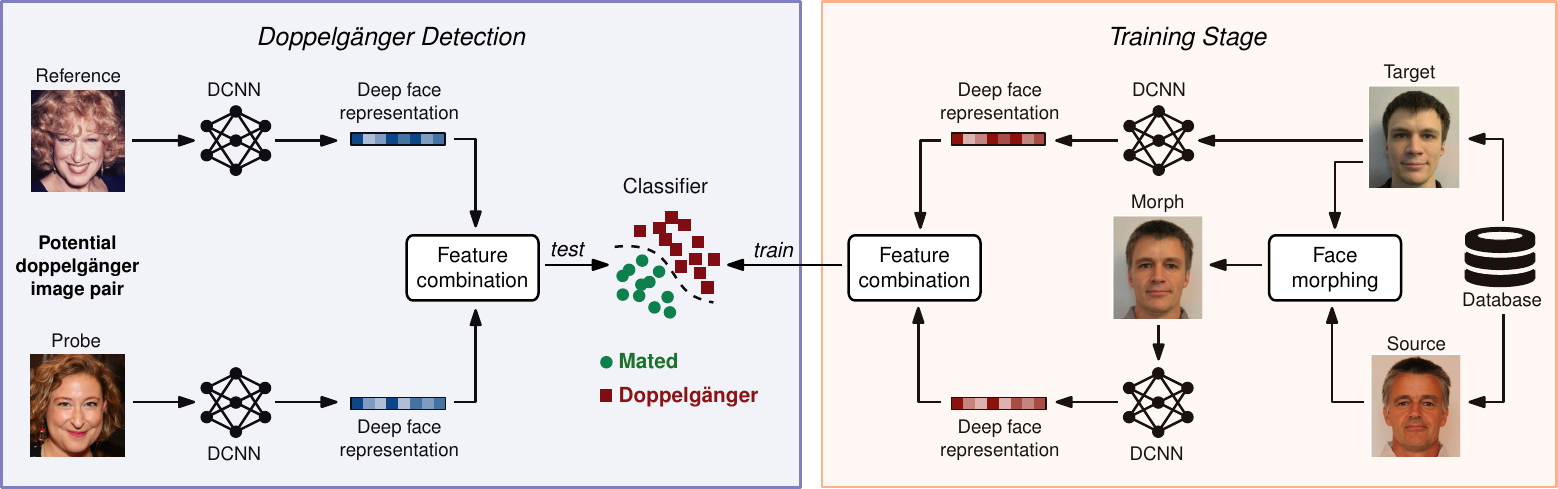}
\caption{Overview of the proposed doppelgänger detection system.}\label{fig:system}\vspace{-0.2cm}
\end{figure*}

\section{Related Works}\label{sec:related}
An overview of relevant works which investigated the impact of doppelgängers on face recognition and proposed detection methods is given in table~\ref{table:related}. In a preliminary study,  Lamba \etal~\cite{Lamba11} introduced the Look-Alike Face (LAF) database and investigated the ability of humans and automated face recognition to distinguish lookalikes. Their analysis showed that neither humans nor automatic face recognition algorithms were able to correctly detect lookalikes, obtaining error rates close to guessing. As countermeasure to this vulnerability, a facial region-based feature extraction was proposed in   \cite{Lamba11} of which the result is fed to a support vector machine (SVM) to distinguish lookalikes. Moeini \etal~\cite{MOEINI20171} suggested to employ a textural analysis together with 3D reconstruction methods and a classification based on Sparse Representation Classification (SRC) in order to differentiate lookalike face image pairs. On the LAF dataset, a  detection equal error rate of approximately 6\% was reported in \cite{MOEINI20171}. On the same database, inferior results were achieved by the detection scheme of Rustam and Faradina \cite{Rustam_2019}. More recently, Swearingen and Ross \cite{Swearingen20} presented an approach to improve facial identification performance by re-ranking candidate lists using a lookalike disambiguator which is specifically trained to distinguish between lookalike and mated face image pairs. In \cite{Swearingen20}, slight improvements were reported on the challenging low-resolution TinyFace database.

Some works have focused on collecting doppelgänger image pairs which can be used to re-train face recognition techniques or provide a more challenging evaluation scenario. To learn highly discriminative facial representations which should also allow to distinguish doppelgängers, Smirnov \etal~\cite{Smirnov17} refined the mini-batch selection of a general-purpose face recognition model using a list of lookalikes. In \cite{Smirnov17}, it was shown that such a re-training can slightly improve face recognition accuracy. Deng \etal \cite{deng2017fine} introduced the Fine-Grained LFW (FGLFW) dataset, a subset of the Labeled Faces in the Wild (LFW) database, which was selected by human crowdsourcing. A significant decrease in recognition performance on FGLFW (compared to LFW) was reported in \cite{deng2017fine} for several state-of-the-art face recognition techniques. In their Disguised Faces in the Wild (DFW) dataset, Singh \etal~\cite{Singh19} collected facial images which represent challenging face recognition scenarios, including lookalike pairs.  Swearingen and Ross \cite{Swearingen20} automatically derived doppelgänger image pairs  from the TinyFace dataset  through the analysis of face recognition scores. Similar to the LAF and DFW database, Rathgeb \etal~\cite{Rathgeb-Doppelgaenger-BIOSIG-2021} recently introduced the web-collected HDA-DG database consisting of 400 high quality image pairs of doppelgängers (with gender parity). For different face recognition systems it was shown that the HDA-DG dataset is even more challenging than the DFW dataset.

\begin{figure*}[!t]
\vspace{-0.0cm}
\centering
\includegraphics[height= 3cm]{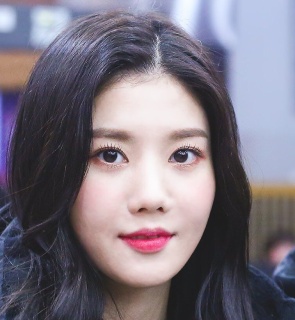} \includegraphics[height= 3cm]{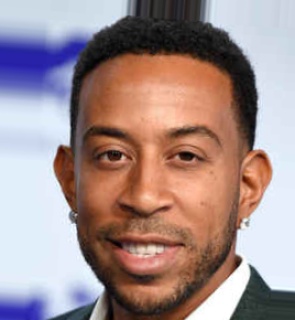}  \includegraphics[height= 3cm]{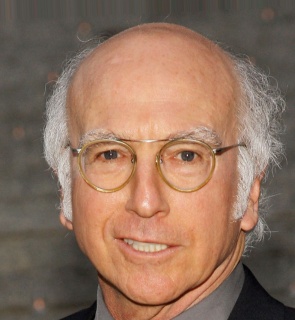}
\includegraphics[height= 3cm]{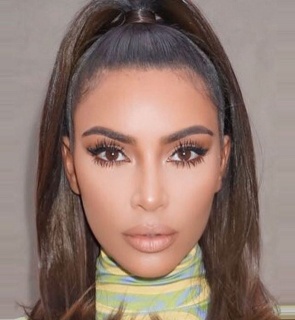}\vspace{0.1cm} \includegraphics[height= 3cm]{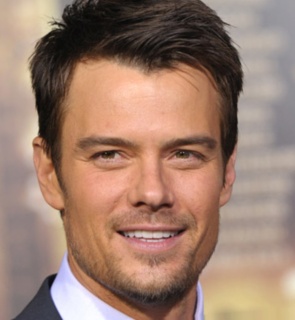} \includegraphics[height= 3cm]{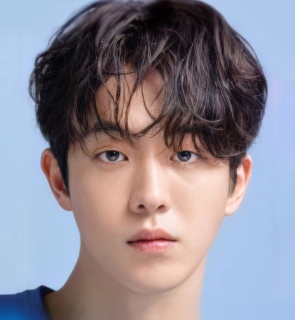}
\includegraphics[height= 3cm]{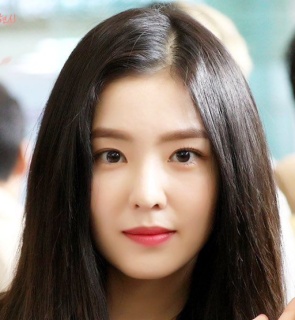} \includegraphics[height= 3cm]{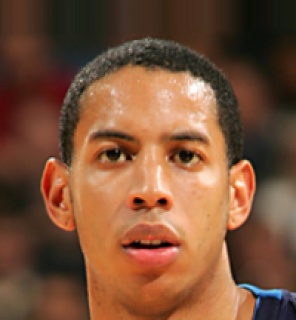} \includegraphics[height= 3cm]{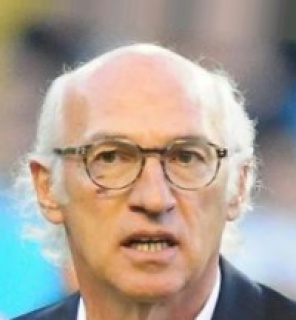}
\includegraphics[height= 3cm]{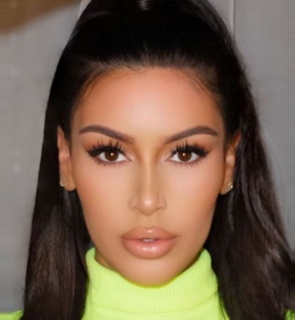} \includegraphics[height= 3cm]{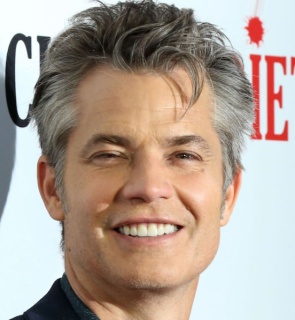} \includegraphics[height= 3cm]{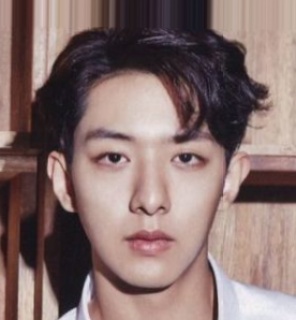} 

\caption{Example doppelgänger image pairs (column-wise) from the HDA Doppelgänger database.}\label{fig:db}\vspace{-0.2cm}
\end{figure*}

Apart from works directly focusing on the impact and detection of doppelgängers on face recognition, there exist several highly related fields of research. Specific efforts have been devoted to differentiate monozygotic twins in the framework of a facial recognition system, \eg through the analysis of facial marks \cite{Srinivas12}. The vulnerability of face recognition schemes to presentation attacks has been investigated by numerous researchers, see \cite{Galbally-Survey-Face-PAD-IEEEAccess-2014,Raghavendra-FacePAD-Survey-2017}. In particular, makeup-based presentation attacks have recently been considered, \eg in \cite{Chen17a}. To launch a makeup presentation attack, an attacker alters his face to obtain the facial appearance of a target subject. Some detection schemes have already been suggested by different researchers, see \cite{Rathgeb-MakeupAttackDetection-ACCESS-2020}. Moreover, the vulnerability of face recognition towards so-called morphing attacks has been exposed in recent years \cite{Ferrara-TheMagicPassport-IJCB-2014,Scherhag-MorphingSurvey-2019,Rathgeb-Rathgeb-HandbookFaceManipulationDetection-2022}. In a morphing attack, face images of two (or more) subjects are combined to a morph such that images of the subjects contributing to the morph will successfully match against it. To detect such digitally manipulated facial images, various techniques have been proposed, see \cite{9115874,Rathgeb-Rathgeb-HandbookFaceManipulationDetection-2022}. It has been shown that in scenarios, where trusted facial imagery is available along with a suspected facial image, a robust and reliable differential detection is feasible, \eg in \cite{Scherhag-FaceMorphingAttacks-TIFS-2020,Cozzolino_2021_ICCV,Ibsen-WIFS-2021}. Similar concepts can be applied to perform an image pair-based detection of doppelgängers.

\section{Doppelgänger Detection}\label{sec:system}
The following subsections describe the key components of the proposed doppelgänger detection system, see figure~\ref{fig:system}. Obviously, doppelgängers cannot be detected from a single image without any target subject. Therefore, a differential detection system is designed which processes an image pair, \ie a reference and a probe image. Differences between facial features extracted from image pairs are learned in a training stage employing a machine learning-based classifier. The following subsections describe the employed extraction of deep face representations and the machine learning-based classification (section~\ref{sec:feature}) as well as the generation of doppelgänger image pairs for training purposes (section~\ref{sec:training}).

\subsection{Feature Extraction and Classification}\label{sec:feature}
Given a reference and a probe image, faces are detected, normalised, and deep face representations are extracted from both images using a deep convolutional neural network (DCNN) of a state-of-the-art face recognition algorithm (see section~\ref{sec:db}). Deep face recognition systems leverage very large databases of face images to learn rich and compact representations of faces. It is expected that differences between doppelgängers will, to a certain extent, also be reflected in extracted deep face representations, \ie the outputs of the DCNN on the last layer. However, this hypothesis is difficult to verify since deep face representations are not human interpretable.


Basically, it would be feasible to train a DCNN from scratch or to apply transfer learning and re-train a pre-trained deep face recognition network to detect doppelgängers. However, the high complexity of the model, represented by the large number of weights in a DCNN, requires a huge amount of training data. Even if only the last layers are re-trained, the limited number of training images (and much lower number of subjects) in used databases can easily result in overfitting to the characteristics of the training set.
 
During doppelgänger classification, a pair of deep face representations extracted from a reference and probe face image are combined by estimating their difference vector. Specifically, an element-wise subtraction of feature vectors is performed. It is expected that differences in certain elements of difference vectors indicate doppelgängers. In the training stage, difference vectors are extracted and a Support Vector Machine (SVM) with a Radial Basis Function (RBF) kernel is trained to distinguish mated comparisons and doppelgängers.

\subsection{Training Data Generation}\label{sec:training}
Due to the fact that there exists no sufficiently large publicly available face database containing high quality doppelgänger image pairs, we automatically generate a database of \emph{synthetic doppelgängers}. Note that the doppelgänger dataset used in \cite{Swearingen20} consists of very low resolution images and the impersonation subset of the DFW database, collected in \cite{Singh19}, contains impostors which exhibit a rather low chance for being false matched against a target subject \cite{Rathgeb-Doppelgaenger-BIOSIG-2021}. Therefore, face morphing techniques are applied to create doppelgänger image pairs from an image pair consisting of a \emph{target} and \emph{source}. The goal of the face morphing is to create an artificial image by changing the facial shape and texture of the target image according to a chosen source image. The resulting morph together with the target image represents the synthetic doppelgänger image pair, see figure~\ref{fig:system} (right part). In fact, depending on the chosen parameters and the applied morphing technique, the source together with the morph could also be considered as a doppelgänger image pair  and, thus, be used for training purposes. Morphing usually consists of two main processes:
\begin{enumerate}
\item \emph{Image warping}: doppelgängers may not have identical but similar facial shape. To simulate similarities in facial shapes, image warping \cite{Glasbey98} is applied as part of the morphing process. Specifically, facial landmarks of a target  image and the source image are extracted. Subsequently, image warping is applied to the target face image \wrt the landmarks detected in the source image applying some weight. Depending on the chosen weights the resulting morphed face image will then exhibit a mixture of the facial shape of the target and the source image. 
\item \emph{Texture blending}: the facial textures of doppelgängers may as well be similar to a certain extent. This is simulated by alpha blending the source image over the target image. Depending on the chosen alpha value the resulting morph will contain textural features of both, the target and the source image.  
\end{enumerate} 

The aforementioned processing steps are applied to pairs of target and source images. For the source and target images, frontal pose, relatively neutral facial expression (\eg closed mouth), and similar sample quality are assured, \cf figure~\ref{fig:morph_example}. During training, synthetically generated doppelgänger pairs are used together with  mated comparison trails, \ie unaltered pairs of face images of the same subject. Generated doppelgänger pairs are only used in the training stage, \ie the use of synthetically generated images in both training and testing is deliberately avoided in order to prevent the classifier from any sort of overfitting towards potential artefacts introduced by the morphing process.

\section{Experimental Setup}\label{sec:db}
The following subsections describe the software and databases (subsection~\ref{subsec:dbs}), as well as evaluation methodology and metrics (subsection~\ref{subsec:metrics}) used in the experimental evaluations of the proposed doppelgänger   detection system. 

\subsection{Databases and Software}\label{subsec:dbs}

The HDA-DG database introduced in \cite{Rathgeb-Doppelgaenger-BIOSIG-2021} is used for the vulnerability analysis and the evaluation of the proposed detection method. This database was collected from the web using search terms like ``lookalikes'' or ``doppelgängers''. A total number of 400 mostly frontal doppelgänger image pairs was collected and manually checked. During the collection, gender parity as well as diversity in other demographic attributes was assured, resulting in 200 male and female image pairs of various age groups and skin colours. Example image pairs of the collected dataset are shown in figure~\ref{fig:db}. Additionally, the DFW dataset introduced in \cite{Singh19} was used during the vulnerability analysis.  Similar to the HDA-DG dataset, the majority of facial images of DFW are of celebrities. In addition to these datasets, mated and non-mated comparison trials were obtained from the FRGCv2 face database \cite{Phillips-2005}. In order to enable a direct comparison between the proposed detection method and previous works, detection experiments are additionally carried out on the LAF database \cite{Lamba11}. For the training of the proposed detection scheme a subset of the FERET face database \cite{Phillips-FERET-IEEE-2000} was used, \ie training and evaluation of the proposed method is performed in a cross-database experiment. For the creation of morphs, the face images of the FERET face database are lexicographically sorted, and then each input image is morphed with one of the next consecutive input image, for which the following two criteria hold: both depicted subjects are of same sex; only one face image of the image pair shows glasses. The latter criteria should avoid artefacts. Each facial image is only used once, \ie for the generation of a single morph. This simple pairing algorithm has been found to be suffciently effective while more sophisticated pairing algorithms have been proposed, \eg in \cite{Roettcher20}. In contrast, the used pairing method is expected to generate morphs from subject pairs exhibiting various degrees of similarity.

A COTS face recognition system is used in the vulnerability analysis which raises the practical relevance. Since the COTS system is closed-source it is only used in the vulnerability analysis. It is also based on deep learning like the majority of state-of-the-art face recognition systems. For each face image pair, the used COTS face system produces similarity scores in the range $[0,1]$.

In the proposed detection system, the \emph{dlib} algorithm \cite{King2009} is applied for face detection and facial landmark extraction. Deep face representations are extracted using the well-known open-source ArcFace \cite{Deng19} system with a pre-trained model r50-am-lfw. The ArcFace algorithm is a widely used DCNN that has been shown to obtain competitive biometric performance on many challenging databases. The extracted feature vectors consist of 512 floating-point values. After the calculation of the differential feature vectors of the probe and target images, a python standard scaler is trained using the feature vectors of the training set. This scaler is then applied to the feature vectors for training as well as testing. Subsequently, the \emph{libsvm} library \cite{libsvm} is used to train SVMs employing standard parameters. Trained SVMs generate a normalised detection score in the range $[0,1]$,  0 meaning a higher probability for being a mated comparison and 1 to be a image pair of doppelgängers.

\begin{figure}[!t]
\centering
	\begin{minipage}{0.3\linewidth}
		\includegraphics[height=3.5cm]{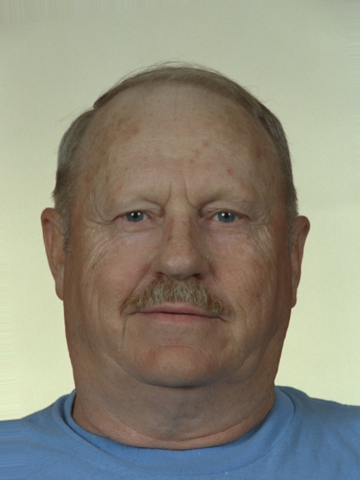}\newline\vspace{-0.5cm}
		\subfigure[Target]{\includegraphics[height=3.5cm]{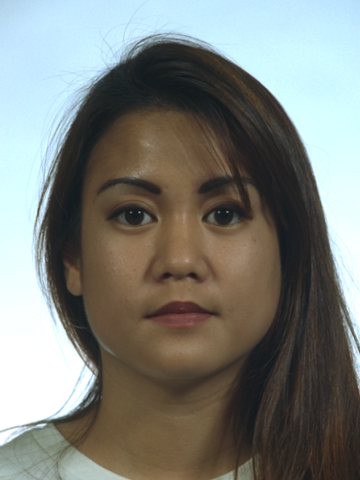} \label{fig:sample_subject1}}
	\end{minipage}
	\begin{minipage}{0.3\linewidth}
		\includegraphics[height=3.5cm]{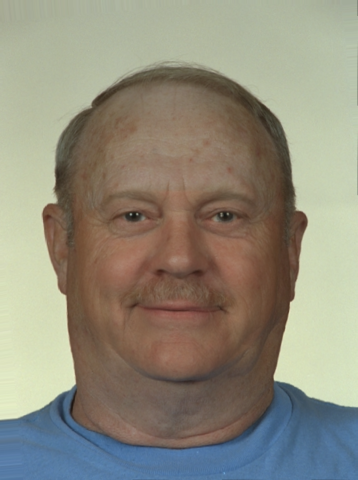}\newline\vspace{-0.5cm}
		\subfigure[Morph]{\includegraphics[height=3.5cm]{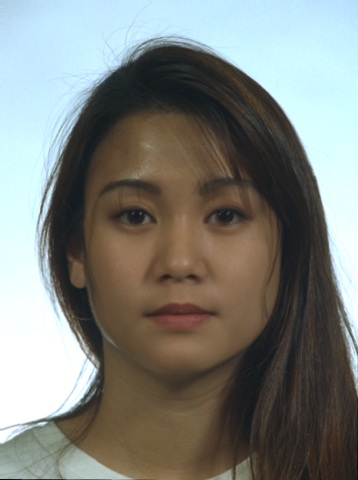} \label{fig:sample_facefusion}}
	\end{minipage}
	\begin{minipage}{0.3\linewidth}
		\includegraphics[height=3.5cm]{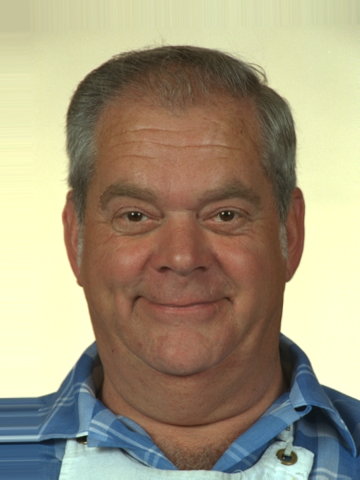}\newline\vspace{-0.5cm}
		\subfigure[Source]{\includegraphics[height=3.5cm]{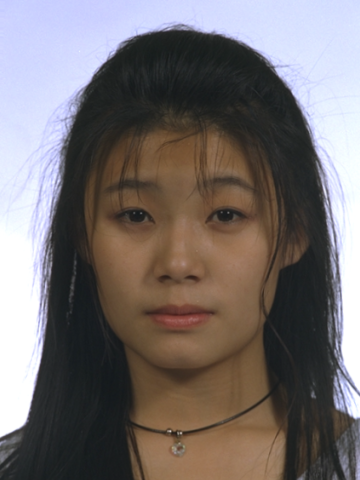} \label{fig:sample_subject2}}
	\end{minipage}	
	\caption{Examples of generated doppelgänger face images. Equal weights of face images were used to create morphs. Face images depicting subjects with similar demographic attributes, e.g., age and ethnicity, generally result in more plausible face morphs. }
	\label{fig:morph_example}
\end{figure}

During training, two different morphing algorithms are employed, see figure~\ref{fig:morph_example}:
\begin{itemize}
	\setlength
	\item FaceFusion\footnote{\url{www.wearemoment.com/FaceFusion/}}, a proprietary morphing algorithm. 
	\item UBO, the morphing tool of University of Bologna, as used \eg in \cite{Ferrara2018}. 
\end{itemize}
It can be seen that for both algorithms the generated morph, \ie the synthetic doppelgänger face image, retains the outer facial region of the target subject. This is reasonable, since doppelgängers usually imitate things like hair style or accessories, \cf figure~\ref{fig:ex}.  For both algorithms fixed blending and warping factors of 0.5 are used, \ie the target and the source image each contribute 50\% to the inner facial region in the doppelgänger image.

\subsection{Evaluation Metrics}\label{subsec:metrics}
The used evaluation metrics conform to the currently applicable international standards for biometric performance and presentation attack detection, \ie ISO/IEC 19795-1:2021 and ISO/IEC 30107-3:2017 \cite{ISO-PerformanceReporting-2021,ISO-IEC-30107-3-PAD-metrics-170227}. As previously mentioned, doppelgängers do not neccessarly represent an attack. Nevertheless, the metrics defined in \cite{ISO-IEC-30107-3-PAD-metrics-170227} are suitable for evaluating the detection performance of the proposed system.\footnote{ISO/IEC 19795-1:2021 lists ``zero effort impostor attempts'' as conformant human PAI.} Specifically, following metrics are reported:

\begin{table}[!t]
\centering
\caption{Number of comparisons for  HDA-DG and DFW  databases.}
\label{table:datasets}
\resizebox{0.65\linewidth}{!}{
\begin{tabular}{lrr}
\toprule
\multicolumn{1}{l}{\multirow{1}{*}{\textbf{Comparisons}}} & \multicolumn{1}{c}{\textbf{HDA-DG}} & \multicolumn{1}{c}{\textbf{DFW}} \\ 
\midrule
Doppelgänger   & 389   & 4,305   \\ 
Mated    & 6,375  & 893 \\ 
Non-mated  & 3,664,320  & 496,506 \\ 
\bottomrule
\end{tabular}
}
\end{table}

\begin{figure*}[!t]
\vspace{-0.0cm}
\centering
\subfigure[PDFs]{\includegraphics[width=0.45\textwidth]{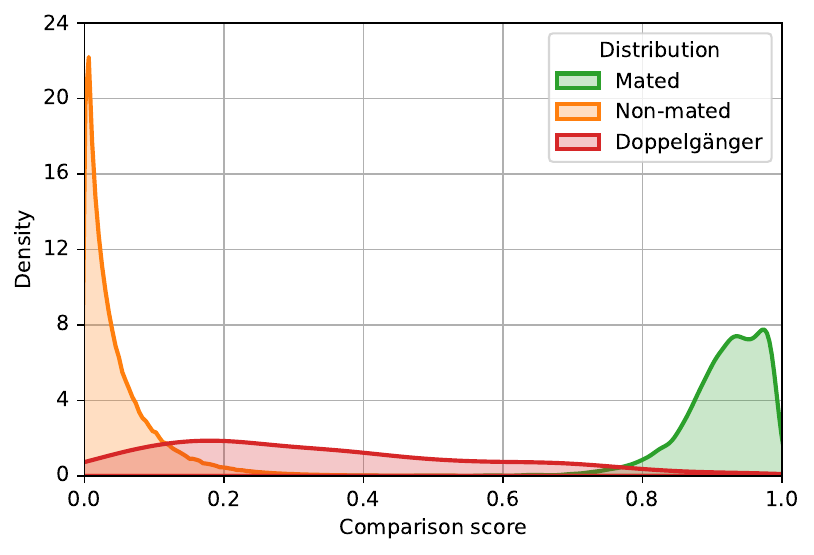}
}
\subfigure[Boxplots]{\includegraphics[width=0.49\textwidth]{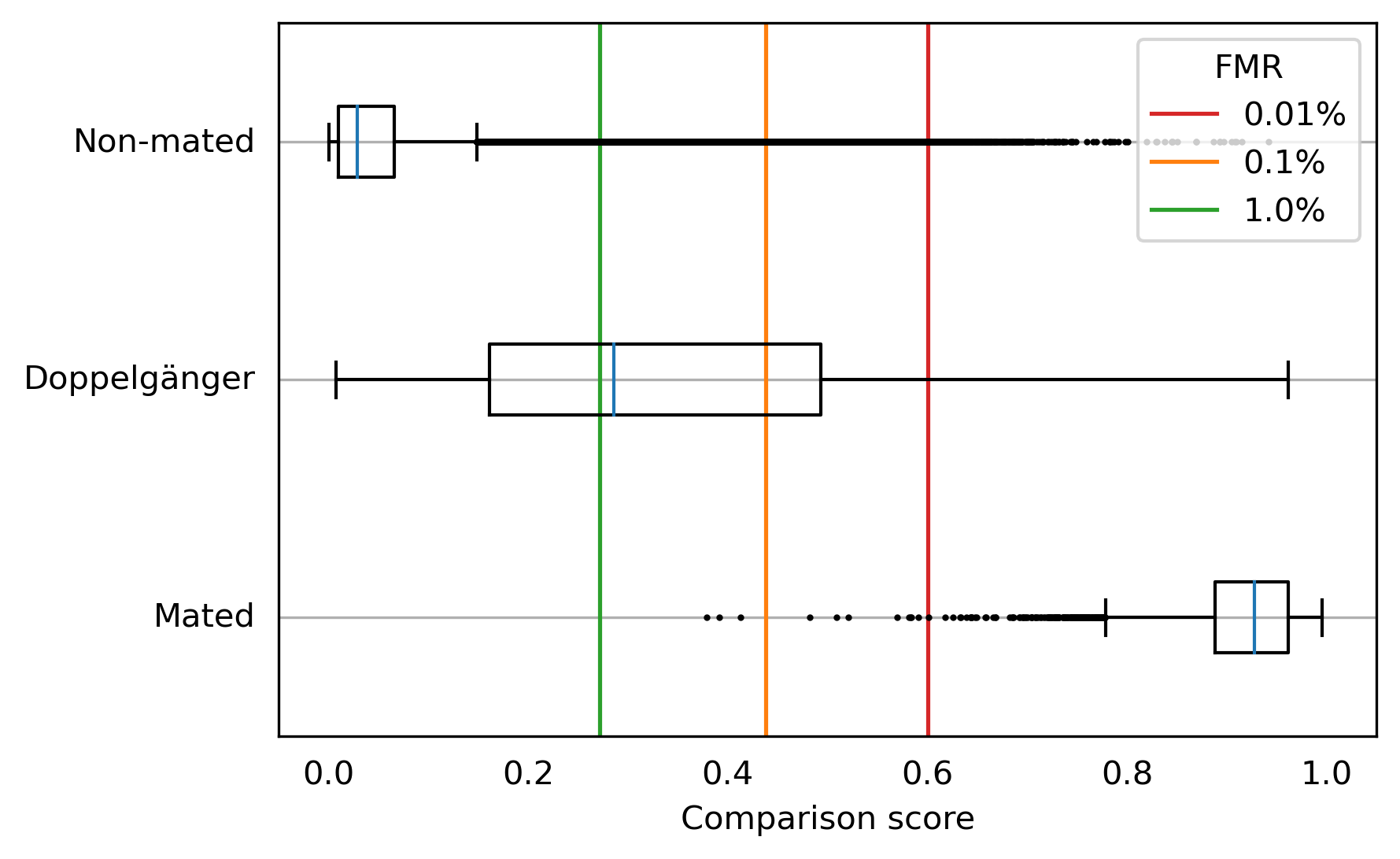}
}
\caption{Probability density functions of comparison scores and boxplots for the HDA-DG database.}\label{fig:score_dists}\vspace{-0.2cm}
\end{figure*}

\begin{figure*}[!t]
\vspace{-0.0cm}
\centering
\subfigure[PDFs]{\includegraphics[width=0.45\textwidth]{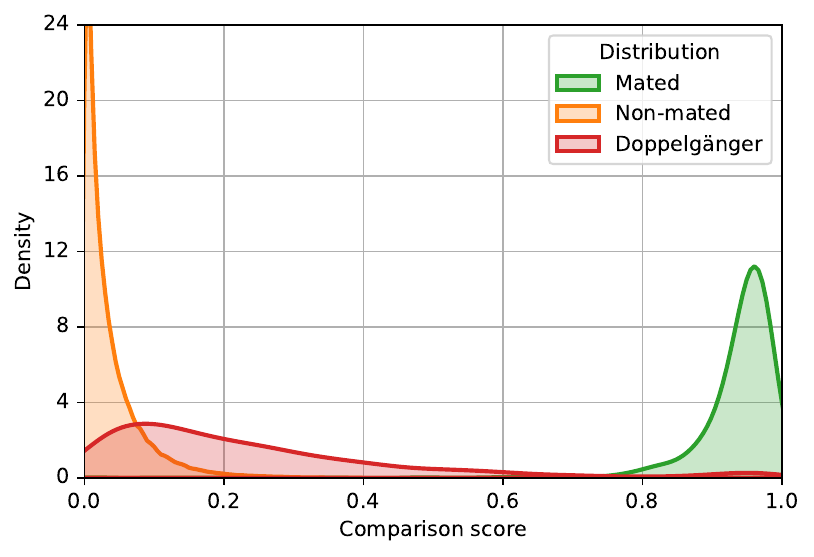}
}
\subfigure[Boxplots]{\includegraphics[width=0.49\textwidth]{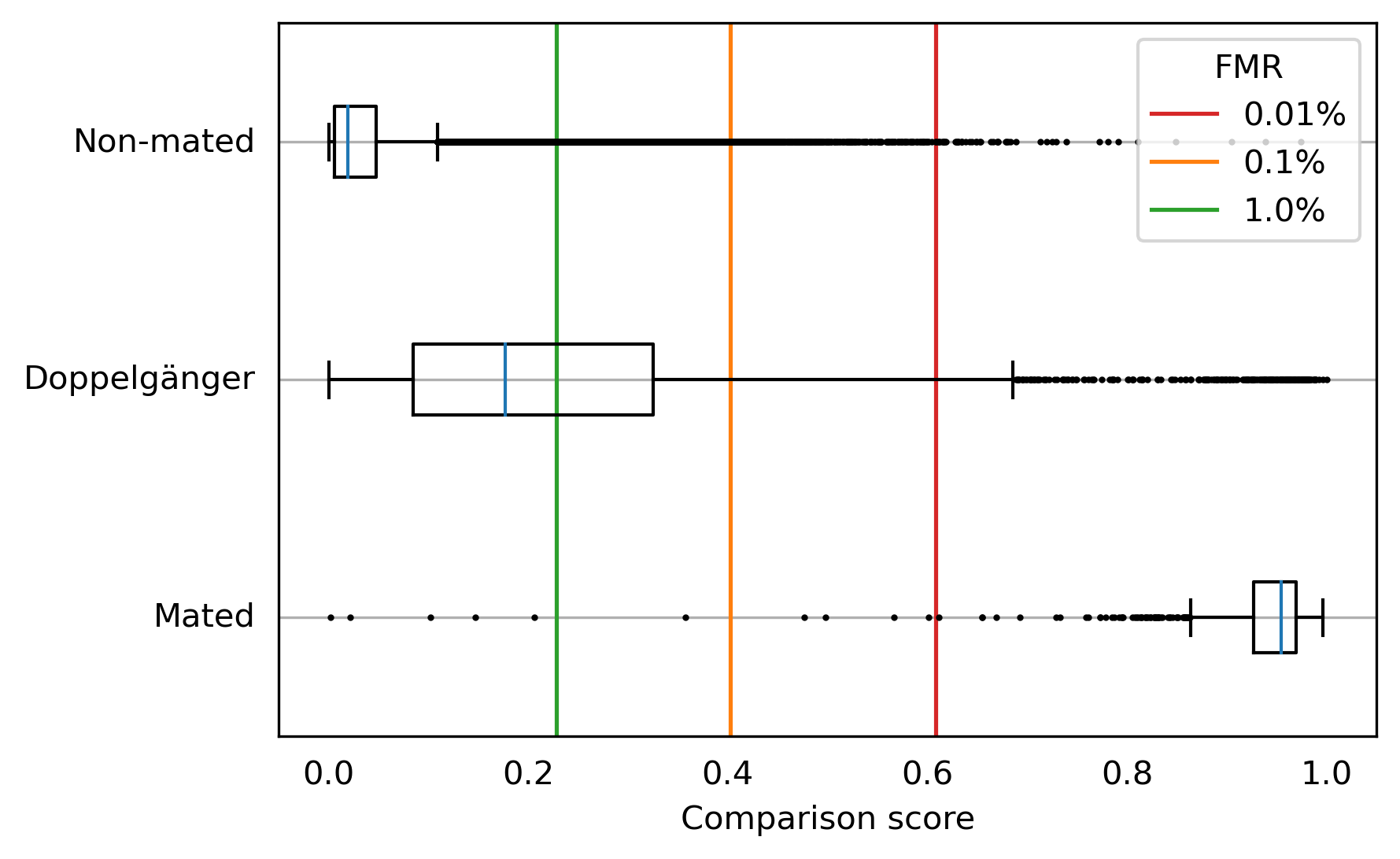}
}
\caption{Probability density functions of comparison scores and boxplots for the DFW database.}\label{fig:score_boxplots}\vspace{-0.0cm}
\end{figure*}

\begin{itemize}
\item \emph{Biometric performance:} the False Non-Match Rate (FNMR) and False Match Rate (FMR) denote the proportion of falsely classified mated, \ie genuine, and non-mated, \ie impostor, attempts in a biometric verification scenario. 
\item \emph{Vulnerability assessment:} the Impostor Attack Presentation Match Rate (IAPMR) \cite{ISO-IEC-30107-3-PAD-metrics-170227} defines the proportion of attack presentations using the same PAI species in which the target reference is matched. In this work, a zero-effort impostor attack of a doppelgänger is interpreted as attack.
\item \emph{Detection performance:} the Attack Presentation Classification Error Rate (APCER) is defined as the proportion of attack presentations using the same PAI species incorrectly classified as bona fide presentations in a specific scenario. The Bona Fide Presentation Classification Error Rate (BPCER) is defined as the proportion of bona fide presentations incorrectly classified as PAs in a specific scenario. Further, as suggested in the aforementioned standard, the BPCER10 and BPCER20 represent the operation points ensuring a security level of the system at an APCER of 10\% and 5\%, respectively. Additionally, the Detection Equal-Error Rate (D-EER) is reported.
\end{itemize}

\section{Results}\label{sec:experiments}
The impact of doppelgängers on face recognition performance, \ie vulnerability analysis, is presented in the following subsection (subsection \ref{sec:impact}). Subsequently, results obtained for the proposed doppelgänger detection system on the HDA-DG database are reported (subsection \ref{sec:det_perf}). A comparison of the proposed method with previous works on the LAF database is presented in (subsection~\ref{sec:comparison}).

\subsection{Impact on Face Recognition}\label{sec:impact}

\begin{table*}[!t]
\centering
\caption{Descriptive statistics of the HDA-DG and the DFW databases.}\label{table:descriptive_statistics}
\resizebox{0.95\textwidth}{!}{
\begin{tabular}{lrrrrrrrrrrrrr}
\toprule
 \multirow{2}{*}{\textbf{Comparisons}} & \multicolumn{6}{c}{\textbf{HDA-DG}} & & \multicolumn{6}{c}{\textbf{DFW}}  \\ \cmidrule(r){2-8} \cmidrule(l){9-14} 
                        & Mean & Min. & Max. & Std. dev. & Skew. & Ex. kurt. & & Mean & Min. & Max. & Std. dev. & Skew. & Ex. kurt.  \\ \midrule
 Doppelgänger                                    & 0.34  & 0.01 & 0.96 & 0.23   & 0.71 & -0.32  &  & 0.24 & 0.00 & 0.99  & 0.22  & 1.58 & 2.41    \\ 
 Mated                                        & 0.92  & 0.37 & 0.99 & 0.06    & -1.68 & 6.13  & & 0.93  & 0.00 & 0.99 & 0.09   & -6.59 & 57.19   \\ 
 Non-mated                                       & 0.05  & 0.00 & 0.91 & 0.06  & 2.63 & 9.88    & & 0.04  & 0.00 & 0.97 & 0.05  & 3.23 & 17.85    \\ \bottomrule
\end{tabular}
}
\end{table*}

\begin{figure}[!t]
\vspace{-0.0cm}
\centering
\setlength\tabcolsep{2pt}
\begin{tabular}{ccc}
\includegraphics[height= 3cm]{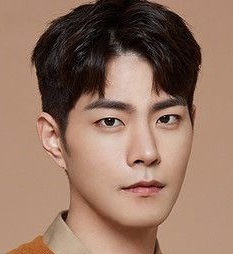}&\includegraphics[height= 3cm]{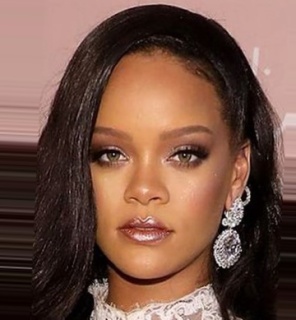}& \includegraphics[height= 3cm]{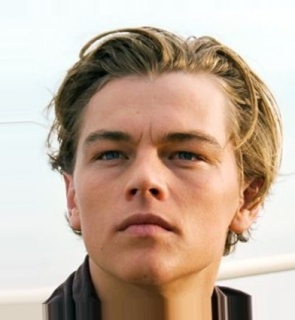} \\
\includegraphics[height= 3cm]{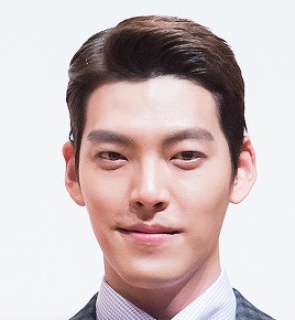}&\includegraphics[height= 3cm]{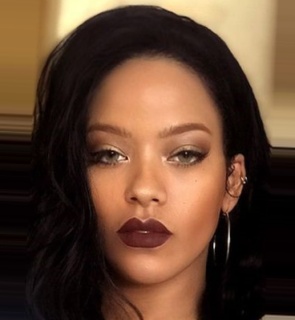}& \includegraphics[height= 3cm]{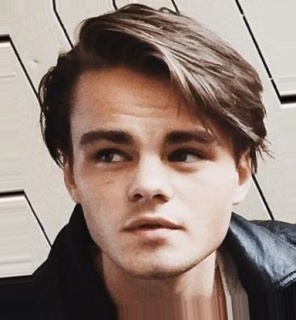}\\
\textsf{0.81} & \textsf{0.73} & \textsf{0.78} \\
\end{tabular}\vspace{-0.2cm}
\caption{Example doppelgänger image pairs (column-wise) and with high comparison scores.}\label{fig:highscores}\vspace{-0.0cm}
\end{figure}

Table~\ref{table:datasets} summarises the number of comparisons (doppelgänger, mated, and non-mated) for the HDA-DG and the DFW database. Note that, doppelgänger comparisons represent a special form of non-mated comparisons. However, non-mated and doppelgänger image pairs are disjoint sets in the conducted experiments. In some cases the COTS system failed to extract the deep face representations. Table~\ref{table:descriptive_statistics} lists descriptive statistics of the resulting score distributions. Score distributions and corresponding boxplots for the HDA-DG and the DFW database are depicted in in figure~\ref{fig:score_dists} and figure~\ref{fig:score_boxplots}, respectively. Here, skewness and excess kurtosis involve the tails of the distributions. Skewness is a measure of a symmetry of the distributions. Kurtosis is a measure of the combined weight of the tails relative to the rest of the distribution whereas excess kurtosis is the difference between the kurtosis of a distribution and that of a normal distribution. It can be observed that the comparison scores obtained from lookalike face image pairs are generally higher compared to the non-mated scores. Further, it can be seen that the doppelgängers of the HDA-DG database yield higher comparison scores than those of the DFW database. Moreover, we observe that doppelgänger score distributions exhibit high standard deviations and longer tails. That is, some doppelgänger image pairs yield very high comparison scores while the overall distribution is skewed towards the non-mated score distribution. This is further pronounced in the corresponding comparison score boxplots which additionally include decision thresholds obtained from the FMRs. Examples of doppelgängers achieving high comparison scores are shown in figure~\ref{fig:highscores}.

Table~\ref{table:perf} summarises the performance obtained on both databases in the absence/ presence of lookalikes. Without doppelgängers, it can be observed that the COTS face recognition system obtains competitive recognition performances on both datasets (across the considered, practically relevant~\cite{EU-Frontex-BestPracticeABC-2015}, decision thresholds). With doppelgängers, the IAPMRs, \ie success chances for doppelgängers, at corresponding decision thresholds are considerably high. For a conservative decision threshold, \ie FMR of 0.01\%, IAPMR of approximately 17\% is achieved on the HDA-DG  database. As expected, based on the analysis of the score distributions, IAPMRs on the DFW database are a bit lower -- 6.8\%. For more liberal decision thresholds, \eg FMR of 0.1\% or 1\%, IAPMRs quickly raise above approximately 29\% to 52\% for the HDA-DG database and approximately 17\% to 39\% on the DFW database. These IAPMR values show that the employed face recognition system is not capable of reliably distinguishing lookalikes. On both datasets, the obtained IAPMR values are significantly higher than the FMRs expected for random non-mated comparisons. In \cite{Rathgeb-Doppelgaenger-BIOSIG-2021}, similar results have additionally been reported for a state-of-the-art open-source face recognition system.

Finally, it is important to note that high similarity scores may not only be produced by face image pairs which exhibit high visual similarity (in terms of human perception). Like the employed COTS system, the vast majority of current face recognition algorithms is based on deep learning techniques. It is well known that such algorithms often lack explainability and produce counterintuitive results. That is, a face recognition system may falsely return high similarity scores for visually different face images. Similarly, a face recognition system may return low similarity scores for highly similar face images, which can be caused by digital manipulations. Such effects have for instance been shown for so-called master faces \cite{Shmelkin} or adversarial samples \cite{Deb}.

\subsection{Detection Performance}\label{sec:det_perf}

Based on the above observations, the detection algorithm is evaluated on the HDA-DG database as this is more challenging than the DFW impersonation subset. To maintain a clear separation between training and test data, the training of the doppelgänger detection method is carried out using  FERET database using 529 bona fide, \ie mated, and 529 morphed images (per morphing algorithm).  These images are paired with probe images resulting in 786 mated and 786 doppelgänger image pairs.

\begin{figure}[!t]
\vspace{-0.0cm}
\centering
\includegraphics[width=0.8\linewidth]{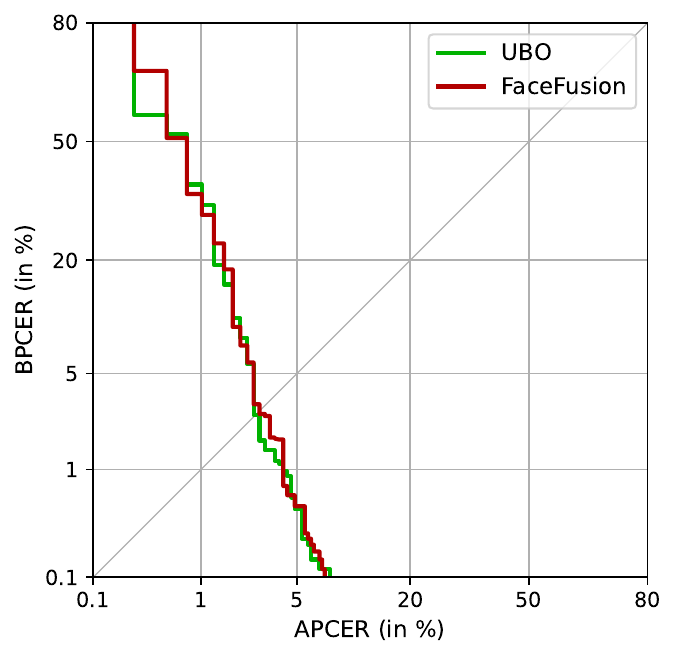}
\caption{DET curve of the detection method on the HDA-DG database.}\label{fig:det}\vspace{-0.2cm}
\end{figure}

Detection accuracies obtained on the HDA-DG database for using different morphing algorithms during training of the doppelgänger detection system are summarised in table~\ref{tab:detectionlaf}.  Corresponding DET curves are depicted in figure~\ref{fig:det}. It can be observed that very similar results are obtained for both morphing algorithms. Using the UBO-Morpher and the FaceFusion morphing algorithms, D-EERs of 2.57\% and 2.79\% are achieved for the task of distinguishing mated comparisons from doppelängers.

These results confirm the effectiveness of the proposed detection system. Moreover, it is important to note that measuring the mere distance between deep face representations without any training does result in worse detection rates. Precisely, a D-EERs of 3.71\% is achieved for solely estimating the Euclidean distance between deep face representations of the ArcFace feature extractor. Hence, the proposed detection method significantly improves upon the underlying face recognition system. This implies that the proposed method learns certain patterns that are useful in separating doppelgängers from mated comparisons. 

\begin{table}[!t]
\centering
\caption{Performance rates and chances of false matches for the HDA-DG and the DFW  databases.}
\label{table:perf}
\begin{tabular}{lccc}
\toprule
\multirow{2}{*}{\textbf{Database}} & \multicolumn{3}{c}{\textbf{FNMR / IAPMR at FMR of}} \\
\cmidrule{2-4}
& 1.00\% & 0.10\% & 0.01\% \\
\midrule
HDA-DG & 0.00\% / 52.44\% &  0.05\% / 29.82\% & 0.17\% / 17.22\%\\
DFW & 0.56\% / 39.70\% & 0.67\% / 17.12\% & 1.12\% / 6.85\%\\
\bottomrule
\end{tabular}

\end{table}

\begin{table}[!t]
\centering
\caption{Detection rates on the HDA-DG and LAF database.}\label{tab:detectionlaf}
\resizebox{0.99\linewidth}{!}{
\begin{tabular}{llrrr}
\toprule
\textbf{Database} & \textbf{Morphing Algorithm} & \textbf{D-EER} & \textbf{BPCER10} & \textbf{BPCER20} \\
\midrule
\multirow{2}{*}{HDA-DG} & UBO & 2.57\% & 0.03\% & 0.45\% \\
& FaceFusion & 2.79\% & 0.06\% & 0.48\% \\\midrule
\multirow{2}{*}{LAF} & UBO & 2.66\% & 0.20\% & 1.53\% \\
& FaceFusion & 2.98\% & 0.41\% & 1.43\% \\
\bottomrule
\end{tabular}

}
\end{table}

\begin{table*}[!t]
\centering
\caption{Comparison with other works on the LAF database.}\label{tab:other}
\resizebox{0.6\linewidth}{!}{
\begin{tabular}{llr}
\toprule
\textbf{Approach} & \textbf{Description} &\textbf{D-EER} \\
\midrule
Lamba~\etal \cite{Lamba11} & 2D log polar Gabor transform with SVM & 41.00\% \\
Rustam and Faradina \cite{Rustam_2019} & Fuzzy Kernel C-Means & 26.00\% \\
Moeini \etal~\cite{MOEINI20171} & Texture and 3D reconstruction with SRC & 5.89\% \\ \midrule
\rowcolor{gray!15}This work & Deep face representations & 2.66\% \\
\bottomrule
\end{tabular}

}\vspace{-0.2cm}
\end{table*}

\subsection{Comparison with other Methods}\label{sec:comparison}

\begin{figure}[!t]
\vspace{-0.0cm}
\centering
\includegraphics[width=0.8\linewidth]{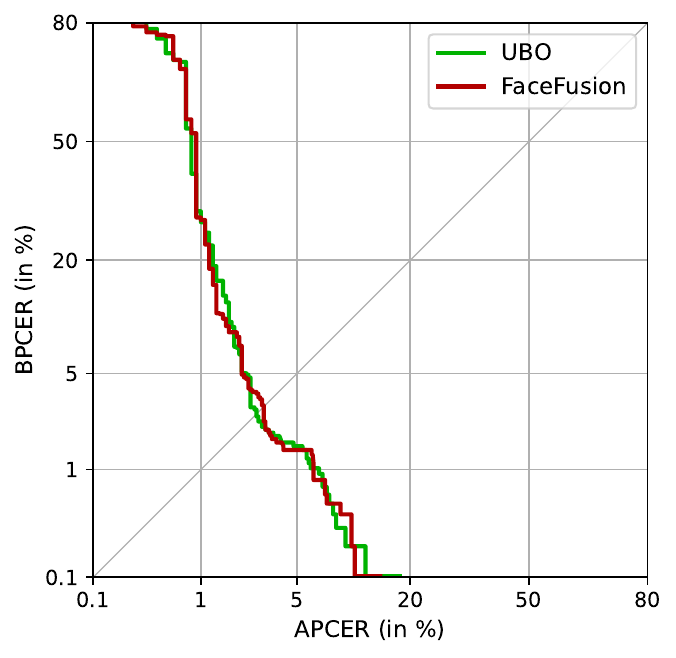}
\caption{DET curve of the detection method on the LAF database.}\label{fig:det2}\vspace{-0.2cm}
\end{figure}

In order to directly compare the proposed method to other published works, it was also evaluated on the LAF database. Obtained results are summarised in table~\ref{tab:detectionlaf} and the corresponding DET curves are shown in figure~\ref{fig:det2}. As can be observed, D-EERs achieved on the LAF are similar to the ones on HDA-DG which indicates that the proposed method generalises well. In addition, table~\ref{tab:other} provides a comparison of the best configuration of the presented approach against previous works in terms of D-EER. It can be seen, that our method significantly outperforms other works for the task of distinguishing doppelgängers from mated comparison trails.

\section{Conclusion}\label{sec:conclusion}
Many face recognition evaluation protocols randomly pair face images to obtain non-mated comparison trails. Obtained non-mated comparison score distribution may then be used to set up decision thresholds at fixed FMRs. It may be concluded that FMRs (and decision thresholds) obtained in such a way overestimate the security of the underlying face recognition system. Furthermore, one may reasonably argue that zero-effort impostor attempts are less likely to be launched by potential attackers that look very different from the attacked target subject. In this work, we assessed the impact of doppelgängers on a COTS face recognition system using the HDA-DG and DFW datasets. It is shown that doppelgänger image pairs achieve very high similarity scores resulting in a significant increase of false match rates.

In addition, we proposed a doppelgänger detection system which analyses differences in deep face representations extracted from a pair of reference and probe face images. Detection scores were obtained from SVM-based classifiers which have been trained to distinguish difference vectors from a training set of mated comparisons and synthetically generated doppelgängers. In experiments on the HDA-DG face database, the proposed doppelgänger detection system was shown to achieve encouraging D-EERs below 2.7\% applying the ArcFace algorithm for the extraction of deep face representations. This performance is maintained on the LAF database where the proposed detection system significantly outperforms previous works.

\section*{Acknowledgements}
\label{sec:acknowledgements}
This research work has been funded by the German Federal Ministry of Education and Research and the Hessian Ministry of Higher Education, Research, Science and the Arts within their joint support of the National Research Center for Applied Cybersecurity ATHENE.

{\small
\bibliographystyle{IEEEtran}
\bibliography{references}
}

\end{document}